\documentclass{article}

\usepackage{arxiv}

\usepackage[utf8]{inputenc} \usepackage[T1]{fontenc}    \PassOptionsToPackage{hyphens}{url}
\usepackage[breaklinks,backref=page,hidelinks]{hyperref}

\renewcommand*{\backref}[1]{}
\renewcommand*{\backrefalt}[4]{\ifcase #1 No citations.\or
(Cited on page: #4).\else
(Cited on pages: #4).\fi }

\usepackage{url}            \usepackage{booktabs}       \usepackage{amsfonts}       \usepackage{nicefrac}    
   \usepackage{microtype}      \usepackage{lipsum}              \usepackage{graphicx}
\usepackage[numbers,sort&compress]{natbib}
\usepackage{doi}
\usepackage{float}
\usepackage{multirow}
\usepackage{lineno}
\usepackage{xcolor}

\usepackage{cleveref}

\usepackage{wrapfig}

\usepackage{authblk}

\usepackage[justification=centering]{caption}

\usepackage[inline]{enumitem}

\usepackage{bbding}

\usepackage[flushleft]{threeparttable}

\usepackage{adjustbox}

\usepackage{listings}

\definecolor{gray}{gray}{0.5}
\colorlet{commentcolour}{green!50!black}

\colorlet{stringcolour}{red!60!black}
\colorlet{keywordcolour}{magenta!90!black}
\colorlet{exceptioncolour}{yellow!50!red}
\colorlet{commandcolour}{blue!60!black}
\colorlet{numpycolour}{blue!60!green}
\colorlet{literatecolour}{magenta!90!black}
\colorlet{promptcolour}{green!50!black}
\colorlet{specmethodcolour}{violet}

\newcommand*{\literatecolour}{\textcolor{literatecolour}}

\newcommand*{\pythonprompt}{\textcolor{promptcolour}{{>}{>}{>}}}

\definecolor{deepblue}{rgb}{0,0,0.5}
\definecolor{deepred}{rgb}{0.6,0,0}
\definecolor{deepgreen}{rgb}{0,0.5,0}

\colorlet{punct}{red!60!black}
\definecolor{background}{HTML}{EEEEEE}
\definecolor{delim}{RGB}{20,105,176}
\colorlet{numb}{magenta!60!black}
\definecolor{keyword}{rgb}{0.8,0,0}

\lstdefinestyle{mypython}{
  language=python,
  captionpos=b,
  showtabs=true,
  tab=,
  tabsize=2,
  basicstyle=\ttfamily\footnotesize,
  stringstyle=\color{stringcolour},
  showstringspaces=false,
  alsoletter={1234567890},
  otherkeywords={\%, \}, \{, \&, \|},
  keywordstyle=\color{keywordcolour}\bfseries,
  emph={and,break,class,continue,def,yield,del,elif ,else,except,exec,finally,for,from,global,if,import,in,lambda,not,or,pass,print,raise,return,try,while,assert,with},
  emphstyle=\color{blue}\bfseries,
  emph={[2]True, False, None},
  emphstyle=[2]\color{keywordcolour},
  emph={[3]object,type,isinstance,copy,deepcopy,zip,enumerate,reversed,list,set,len,dict,tuple,xrange,append,execfile,real,imag,reduce,str,repr},
  emphstyle=[3]\color{commandcolour},
  emph={Exception,NameError,IndexError,SyntaxError,TypeError,ValueError,OverflowError,ZeroDivisionError},    
  emphstyle=\color{exceptioncolour}\bfseries,
  morecomment=[s]{"""}{"""},
  commentstyle=\color{commentcolour}\slshape,
  emph={[4]ode, fsolve, sqrt, exp, sin, cos,arctan, arctan2, arccos, pi,  array, norm, solve, dot, arange, isscalar, max, sum, flatten, shape, reshape, find, any, all, abs, plot, linspace, legend, quad, polyval,polyfit, hstack, concatenate,vstack,column_stack,empty,zeros,ones,rand,vander,grid,pcolor,eig,eigs,eigvals,svd,qr,tan,det,logspace,roll,min,mean,cumsum,cumprod,diff,vectorize,lstsq,cla,eye,xlabel,ylabel,squeeze},
  emphstyle=[4]\color{numpycolour},
  emph={[5]__init__,__add__,__mul__,__div__,__sub__,__call__,__getitem__,__setitem__,__eq__,__ne__,__nonzero__,__rmul__,__radd__,__repr__,__str__,__get__,__truediv__,__pow__,__name__,__future__,__all__},
  emphstyle=[5]\color{specmethodcolour},
  emph={[6]assert,yield},
  emphstyle=[6]\color{keywordcolour}\bfseries,
  emph={[7]range},
  emphstyle={[7]\color{keywordcolour}\bfseries},
  literate=*
    {:}{{\literatecolour:}}{1}
    {=}{{\literatecolour=}}{1}
    {-}{{\literatecolour-}}{1}
    {+}{{\literatecolour+}}{1}
    {*}{{\literatecolour*}}{1}
    {**}{{\literatecolour{**}}}2
    {/}{{\literatecolour/}}{1}
    {//}{{\literatecolour{//}}}2
    {!}{{\literatecolour!}}{1}
    {[}{{\literatecolour[}}{1}
    {]}{{\literatecolour]}}{1}
    {<}{{\literatecolour<}}{1}
    {>}{{\literatecolour>}}{1}
    {>>>}{\pythonprompt}{3},
  frame=trbl,
  rulecolor=\color{black!40},
  backgroundcolor=\color{background},
  breakindent=.5\textwidth,frame=single,breaklines=true
}

\lstdefinelanguage{json}{
    basicstyle=\footnotesize\ttfamily,
    numberstyle=\scriptsize,
    numbersep=8pt,
    showstringspaces=false,
    breaklines=true,
    frame=lines,
    backgroundcolor=\color{background},
    string=[s]{"}{"},
    stringstyle=\ttfamily\color{deepgreen},
    morekeywords={true, false},
    keywordstyle=\color{keyword},
    captionpos=b,
    literate=
     *{0}{{{\color{deepblue}0}}}{1}
      {1}{{{\color{deepblue}1}}}{1}
      {2}{{{\color{deepblue}2}}}{1}
      {3}{{{\color{deepblue}3}}}{1}
      {4}{{{\color{deepblue}4}}}{1}
      {5}{{{\color{deepblue}5}}}{1}
      {6}{{{\color{deepblue}6}}}{1}
      {7}{{{\color{deepblue}7}}}{1}
      {8}{{{\color{deepblue}8}}}{1}
      {9}{{{\color{deepblue}9}}}{1}
{\{}{{{\color{delim}{\{}}}}{1}
      {\}}{{{\color{delim}{\}}}}}{1}
      {[}{{{\color{delim}{[}}}}{1}
      {]}{{{\color{delim}{]}}}}{1},
}
\title{SC2EGSet: StarCraft II Esport Replay and Game-state Dataset }

\date{July 28, 2022}

\makeatletter
\newcommand\email[2][]{\newaffiltrue\let\AB@blk@and\AB@pand
      \if\relax#1\relax\def\AB@note{\AB@thenote}\else\def\AB@note{\relax}\setcounter{Maxaffil}{0}\fi
      \begingroup
        \let\protect\@unexpandable@protect
        \def\thanks{\protect\thanks}\def\footnote{\protect\footnote}\@temptokena=\expandafter{\AB@authors}{\def\\{\protect\\\protect\Affilfont}\xdef\AB@temp{#2}}\xdef\AB@authors{\the\@temptokena\AB@las\AB@au@str       
         \protect\\[\affilsep]\protect\Affilfont\AB@temp}\gdef\AB@las{}\gdef\AB@au@str{}{\def\\{, \ignorespaces}\xdef\AB@temp{#2}}\@temptokena=\expandafter{\AB@affillist}\xdef\AB@affillist{\the\@temptokena \AB@affilsep          \AB@affilnote{}\protect\Affilfont\AB@temp}\endgroup
       \let\AB@affilsep\AB@affilsepx
}
\makeatother

\author[1]{\textbf{Andrzej Białecki}\textsuperscript{*,}}
\affil[1]{Warsaw University of Technology}

\author[2]{\textbf{Natalia Jakubowska}}
\affil[2]{SWPS University}

\author[3]{\textbf{Paweł Dobrowolski}}
\affil[3]{Institute of Psychology, Polish Academy of Sciences}
\author[ ]{\textbf{Piotr Białecki}}
\author[ ]{\\\textbf{Leszek Krupiński}}
\author[4]{\textbf{Andrzej Szczap}}
\affil[4]{Adam Mickiewicz University in Poznań}
\author[5]{\textbf{Robert Białecki}}
\author[5]{\textbf{Jan Gajewski}}
\affil[5]{Józef Piłsudski University of Physical Education in Warsaw}

\hypersetup{
pdftitle={SC2EGSet: StarCraft II Esport Replay and Game-state Dataset},
pdfsubject={cs.LG, cs.AI},
pdfauthor={Andrzej Białecki, Natalia Jakubowska, Paweł Dobrowolski, Piotr Białecki, Leszek Krupiński, Andrzej Szczap, Robert Białecki, Jan Gajewski},
pdfkeywords={esport, dataset, StarCraft II, machine learning, artificial intelligence},
}

\begin{document}
\maketitle

\let\thefootnote\relax\footnote{Dataset API Homepage: \href{https://github.com/Kaszanas/SC2_Datasets}{https://github.com/Kaszanas/SC2\_Datasets}}
\let\thefootnote\relax\footnote{\textsuperscript{*} Corresponding author: \url{andrzej.bialecki94@gmail.com}}\let\thefootnote\relax\footnote{\textsuperscript{*} Institutional contact: \url{andrzej.bialecki.dokt@pw.edu.pl}}

\begin{abstract}

As a relatively new form of sport, esports offers unparalleled data availability. Despite the vast amounts of data that are generated by game engines, it can be challenging to extract them and verify their integrity for the purposes of practical and scientific use.

Our work aims to open esports to a broader scientific community by supplying raw and pre-processed files from StarCraft II esports tournaments. These files can be used in statistical and machine learning modeling tasks and related to various laboratory-based measurements (e.g., behavioral tests, brain imaging). We have gathered publicly available game-engine generated "replays" of tournament matches and performed data extraction and cleanup using a low-level application programming interface (API) parser library.

Additionally, we open-sourced and published all the custom tools that were developed in the process of creating our dataset. These tools include PyTorch and PyTorch Lightning API abstractions to load and model the data.

Our dataset contains replays from major and premiere StarCraft II tournaments since 2016. To prepare the dataset, we processed 55 tournament "replaypacks" that contained 17930 files with game-state information. Based on initial investigation of available StarCraft II datasets, we observed that our dataset is the largest publicly available source of StarCraft II esports data upon its publication.

Analysis of the extracted data holds promise for further Artificial Intelligence (AI), Machine Learning (ML), psychological, Human-Computer Interaction (HCI), and sports-related studies in a variety of supervised and self-supervised tasks.

\end{abstract}


\section{Introduction}
\label{sec:Introduction}

Electronic sports (esports) are a relatively new and exciting multidisciplinary field of study \cite{Reitman2019,Chiu2021}.
There are multiple groups of stakeholders involved in the business of esports \cite{Scholz2019}.
The application of analytics to sports aims to optimize training and competition performance. New training methods are derived from an ever increasing pool of data and research aimed at generating actionable insights \citep{Pustisek2019,Giblin2016,Baerg2017,Chen2021,Rajsp2020,Kos2018}. Rule changes in sports come at varying time intervals and frequently with unpredictable effects on their dynamics. It is especially relevant to share esports data to assess changes in game design and their impact on professional players, as such changes can 
occur more rapidly due to the (yet) relatively unstrctured nature of esports competition \cite{ElNasr2013,Su2021}.

Advancements in Artificial Intelligence (AI) and Machine Learning (ML) have shown that Reinforcement Learning (RL) agents are capable of outmatching human players in many different types of games \cite{Vinyals2019,Jaderberg2019,Silver2018,Berner2019}.
Psychological research on neuroplasticity has also shown the great potential of video games to induce structural brain adaptation as a result of experience \cite{Kowalczyk2018}. Further, previous studies have shown that playing video games can enhance cognitive functioining in a wide range of domains, including perceptual, attentional and spatial ability \cite{Green2003,Green2012}. Data obtained from esports titles -- including those gathered from high-level tournament performance -- may provide a path to improving the quality and reproducibility of research in this field, especially in contrast to the more variable data that is collected in laboratories and in less competitive settings. A lower technical overhead and more data available for modeling could assist further research in these areas \cite{Alfonso2017,Ghasemaghaei2019,Zuiderwijk2019}.

The sparsity and methodological diversity of research on this topic remain as roadblocks in the study of how 
video games can affect mental functioning. Some scholars recommended further research on esports as a potential path forward \cite{Reitman2019}.
Despite the digital nature of esports -- which are their greatest asset with respect to data gathering -- there seems to be a lack of high-quality pre-processed data published for scientific and practical use. The goal of our work is to mitigate this issue by publishing datasets containing StarCraft II replays and pre-processed data from esports events, classified as "Premiere" and "Major" by Liquipedia in the timeframe from 2016 until 2022 \cite{URLLiquipedia2010}.

A brief summary of the contributions stemming from this work is as follows:
\begin{enumerate*}[label=(\arabic*)]
    \item The development of a set of four tools to work with StarCraft II data;
    \item The collected esports data from various public sources;
    \item The publication of a collection of raw replays after brief pre-processing \cite{BialeckiSC2ReSet2021};
    \item The processing of raw data with our toolset and publication as a dataset \cite{BialeckiEGSetDataset};
    \item and the preparation of an official API to interact with our data using PyTorch and PyTorch Lightning for ease of experimentation in further research \cite{BialeckiDatasetAPI}.
\end{enumerate*}

\section{Related Work}
\label{sec:RelatedWork}

While reviewing StarCraft II related sources, we were able to find some publicly available datasets made in 2013 ``SkillCraft1'' \cite{BlairDataset2013} and 2017 ``MSC'' \cite{Huikai2017}. These datasets are related to video games and in that regard could be classified as ``gaming'' datasets. However, it is not clear what percentage of games included within these datasets contain actively competing esports players and at what levels of skill. Using the SkillCraft1 dataset, the authors distinguished the level of players based on the data. They proposed a new feature in the form of the Perception-Action Cycle (PAC), which was calculated from the game data. This research can be viewed as the first step toward developing new training methods and analytical 
depth in electronic sports. It provided vital information describing different levels of gameplay and optimization in competitive settings \cite{Thompson2013}. In \autoref{table:DatasetComparison} we present a comparison of these two StarCraft II datasets to our own.

There are existing datasets in other games. Due to the major differences in game implementations, these could not be directly compared to our work. Despite that, such publications build upon a similar idea of sharing gaming or esports data for wider scientific audience and should be mentioned. Out of all related work, STARDATA dataset is notable in that it comes from prior generation of StarCraft game. This dataset seems to be the largest StarCraft: Brood War dataset available \cite{Lin2021}. Moreover, in League of Legends, a multimodal dataset including physiological data is available \cite{SmerdovLoLDataset2021}.

\begin{table}[H]
    \centering
    \caption{StarCraft II dataset comparison}
    \begin{threeparttable}
    \resizebox{\textwidth}{!}{\begin{tabular}{lccccccc}
        \multicolumn{1}{c}{Dataset}         & esports & public & replays available & pre-processed & API available & replays & timespan       \\ \hline
        SC2EGSet                            & \Checkmark     & \Checkmark    & \Checkmark          & \Checkmark  & \Checkmark           & 17895     & 2016-2022      \\
        SkillCraft1 \cite{BlairDataset2013} & \XSolidBrush   & \Checkmark    & \XSolidBrush        & \Checkmark  & \XSolidBrush         & 3395      & ND\tnote{+}    \\
        MSC \cite{Huikai2017}               & \XSolidBrush   & \Checkmark    & \Checkmark\tnote{*} & \Checkmark  & \Checkmark           & 36619     & ND\tnote{+}    \\ \hline
    \end{tabular}}
        \begin{tablenotes}
        \item[*] provided by the game publisher
        \item[+] ND - not disclosed
        \end{tablenotes}
    \end{threeparttable}
    \label{table:DatasetComparison}
\end{table}

Related publications focused on in-game player performance analyses and psychological, technical, mechanical 
or physiological indices. These studies were conducted with use of various video games such as: Overwatch \cite{Braun2017,Glass2020}, League of Legends \cite{Blom2019,Ani2019,Aung2018,Maymin2021,Lee2022}, Dota 2 \cite{Gourdeau2020,Hodge2017,Hodge2019,Cavadenti2016,Pedrassoli2020}, StarCraft \cite{SanchezRuiz2017,Stanescu2021,Norouzzadeh2021}, StarCraft II \cite{Helmke2014,Lee2021,Chan2021,Cavadenti2015,Volz2019}, Heroes of the Storm \cite{Gourdeau2020}, Rocket League \cite{Mathonat2020}, and Counter-Strike: Global Offensive \cite{Khromov2019,Koposov2020,Smerdov2019,Xenopoulos2022,Aditya2021}, among others \cite{Galli2011}. In some cases a comparison between professional and recreational players was conducted.

Most studies did not provide data as a part of their publication. In other publications, the authors used replays that were provided by the game publishers or were publicly available online, which are unsuitable for immediate data modeling tasks without prior pre-processing. The researchers used raw files in MPQ (SC2Replay) format with their custom code when dealing with StarCraft II \cite{URLBlizzardS2ClientProto,Xiangjun2020}. Other studies solved technical problems that are apparent when working with esports data and different sensing technologies, including visualization, but with no publication of data \cite{Bednarek2017,Feitosa2015,Afonso2019,Stepanov2019,Korotin2019}.
Some researchers have attempted to measure tiredness in an undisclosed game via electroencephalography (EEG) 
\cite{Melentev2020}, and player burnout using a multimodal dataset that consisted of EEG, Electromyography (EMG), galvanic skin response (GSR), heart rate (HR), eyetracking, and other physiological measures in esports 
\cite{Smerdov2021}.

\subsection{Game Description}

Many of the related works introduce and communicate the properties of the games that they analyze. In case of StarCraft II, we recommend the following description: ``StarCraft II: Legacy of The Void (SC2) contains various game modes: 1v1, 2v2, 3v3, 4v4, Archon, and Arcade.
The most competitive and esports related mode (1v1) can be classified as a two-person combat, real-time strategy (RTS) game. The goal of the game for each of the competitors is either to destroy all of the structures, 
or to make the opponent resign.'' Moreover, StarCraft II contains multiple matchmaking options: ``Ranked game - Players use a built-in system that selects their opponent based on Matchmaking Rating (MMR) points.       
Unranked game - Players use a built-in system that selects their opponent based on a hidden MMR - such games 
do not affect the position in the official ranking.
Custom game - Players join the lobby (game room), where all game settings are set and the readiness to play is verified by both players - this mode is used in tournament games.
Immediately after the start of the game, players have access to one main structure, which allows for further 
development and production of units.'' \cite{Bialecki2021Determinants}.

\section{Material and Methods}
\label{sec:MaterialAndMethods}

\subsection{Dataset Sources and Properties}
\label{sec:DatasetSources}

The files used in the presented information extraction process were publicly available due to a StarCraft II 
community effort. Tournament organizers for events classified as "Premiere" and "Major" made the replays available immediately after the tournament to share the information with the broader StarCraft II community for research, manual analysis, and in-game improvement. Sources include Liquipedia, Spawning Tool, Reddit, Twitter, and tournament organizer websites. All replay packs required to construct the dataset were searched and downloaded manually from the public domain. The critical properties of the presented dataset are as follows:    

\begin{itemize}
        \item To secure the availability of the raw replays for further research and extraction, the SC2ReSet: StarCraft II Esport Replaypack Set was created \cite{BialeckiSC2ReSet2021}.
        \item The replays were processed under the licenses provided by the game publisher: End User License 
Agreement (EULA), and "\nameref{sec:AILicense}" which is available in \autoref{sec:AILicense} of the supplementary material.
        \item Our dataset was created by using open-source tools that were published with separate digital object identifiers (doi) minted for each of the repositories. These tools are indexed on Zenodo \cite{Bialecki2021Preparator,Bialecki2021MapExtractor,Bialecki2021InfoExtractor}.
        \item We have made available a PyTorch \cite{PyTorch2019} and PyTorch Lightning \cite{PyTorch_Lightning_2019} API for accessing our dataset and performing various analyses. Our API is accessible in the form of 
a GitHub repository, which is available on Zenodo with a separate doi. All of the instructions for accessing 
the data and specific field documentation are published there \cite{BialeckiDatasetAPI}.
        \item The presented dataset is currently the largest that is publicly available, and contains information from prestigeous StarCraft II tournaments.
        \item The dataset can be processed under CC BY NC 4.0 to comply with Blizzard EULA and the aforementioned \nameref{sec:AILicense}.
\end{itemize}

\subsection{Dataset Pre-Processing}
\label{sec:DatasetPreProcessing}

Dataset pre-processing required the use of a custom toolset. Initially, the Python programming language was used to process the directory structure which held additional tournament stage information. We include this information in the dataset in a separate file for each tournament, effectively mapping the initial directory structure onto the resulting unique hashed filenames. Moreover, a custom tool for downloading the maps was used; only the maps that were used within the replays were downloaded \cite{Bialecki2021Preparator}. Finally, to 
ensure proper translation to English map names in the final data structures, a custom C++ tool implementation was used. Information extraction was performed on map files that contained all necessary localization data \cite{Bialecki2021MapExtractor}. The entirety of our processing pipeline is visualized in \autoref{fig:DatasetPipeline}, and additional visualizations are available in the Appendix, \autoref{sec:appPipelineVisualizations}.

\begin{figure}[H]
    \includegraphics[width=\linewidth]{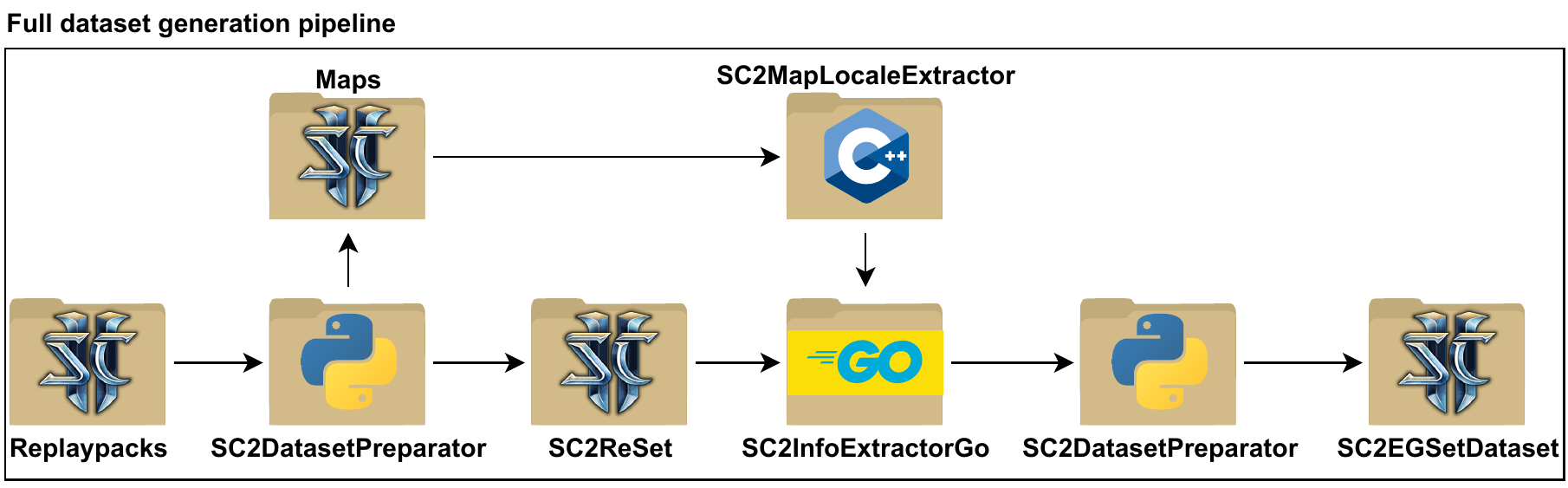}
    \caption{Pre-processing and processing steps of our pipeline that result in SC2ReSet \cite{BialeckiSC2ReSet2021} and SC2EGSetDataset \cite{BialeckiEGSetDataset}. We used a custom data processing toolset including the SC2DatasetPreparator \cite{Bialecki2021Preparator}, SC2MapLocaleExtractor \cite{Bialecki2021MapExtractor}, and SC2InfoExtractorGo \cite{Bialecki2021InfoExtractor}.}
    \label{fig:DatasetPipeline}
\end{figure}

\subsection{Data Processing}
\label{sec:DataProcessing}

Custom software was implemented in the Go programming language (Golang) and built upon authorized and public 
repositories endorsed by the game publisher \cite{URLS2Prot2017}. The tool was used to perform information extraction from files in MPQ format with the SC2Replay extension. Information extraction was performed for each pre-processed directory that corresponded to a single tournament. Depending on the use case, different processing approaches are possible by providing command line arguments \cite{Bialecki2021InfoExtractor}.

\subsection{Data Parsing and Integrity}
\label{sec:DataParsingAndIntegrity}

The parsing capabilities of the tooling were defined with a Golang high-level parser API available on GitHub 
\cite{URLS2Prot2017}. After initial data-structures were obtained, the next step checked the integrity of the data. This was accomplished by comparing information available across different duplicate data structures that corresponded to: the number of players, map name, length of the player list, game version, and Blizzard map boolean (signifying whether a map was published by Blizzard). If a replay parser or custom integrity check 
failed, the replay was omitted.

\subsection{Data Filtering and Restructuring}
\label{sec:DataFilteringAndRestructuring}

Filtering for different game modes was omitted as collected replay files were a part of esports tournament matches. Most often, StarCraft II tournament matches are played in the form of one versus one player combat. Therefore, it was assumed that filtering for the number of players was not required at this step. Custom data structures were created and populated at this stage. This allowed for more control over the processing, summary generation, and final output. Merging data structures containing duplicate information was performed where 
applicable.

\subsection{Summarization and JSON Output to zip archive}
\label{sec:Summarization}

Replay summarization was required in order to provide information that can be accessed without unpacking the 
dataset. Finally, the data was converted from Golang data structures into JavaScript Object Notation (JSON) format, and compressed into a zip archive.

\subsection{Dataset Loading}
\label{sec:DatasetLoading}

Interacting with the dataset is possible via PyTorch \cite{PyTorch2019} and PyTorch Lightning \cite{PyTorch_Lightning_2019} abstractions. Our implementations exposes a few key features:
\begin{enumerate*}[label=(\arabic*)]
        \item Automatic dataset downloading and extraction from Zenodo archive;
        \item Custom validators that filter or verify the integrity of the dataset;
        \item The ability of our abstractions to load and use any other dataset that was pre-processed using 
our toolset.
\end{enumerate*}
The required disk space to succesfully download and extract our dataset is approximately 170 gigabytes. We showcase the example use of our API in \autoref{lst:ExamplePyTorch}. Please note that the API is subject to change and any users should refer to the official documentation for the latest release features and usage information. Additional listing showcasing the use of SC2EGSetDataset is available in the Appendix, \autoref{sec:appDatasetUsage}. Additionally, we include human readable examples of JSON data structures in the Appendix, \autoref{sec:appDatasetStructureExamples}.

\begin{adjustbox}{max width=\textwidth}
\begin{lstlisting}[caption={Example use of the SC2EGSetDataset with PyTorch with a synthetic replaypack prepared for testing.},label={lst:ExamplePyTorch},language=python,style=mypython,firstnumber=1]
from sc2_datasets.torch.sc2_egset_dataset import SC2EGSetDataset
from sc2_datasets.available_replaypacks import (
    EXAMPLE_SYNTHETIC_REPLAYPACKS
)

if __name__ == "__main__":
    # Initialize the dataset:
    sc2_egset_dataset = SC2EGSetDataset(
        unpack_dir="./unpack_dir_path",
        download_dir="./download_dir_path",
        download=True,
        names_urls=EXAMPLE_SYNTHETIC_REPLAYPACKS,
    )

    # Iterate over instances:
    for i in range(len(sc2_egset_dataset)):
        sc2_replay_data = sc2_egset_dataset[i]
\end{lstlisting}
\end{adjustbox}

\section{Dataset Description}
\label{sec:Results}

\begin{figure}[H]
    \centering
    \includegraphics[width=0.8\linewidth]{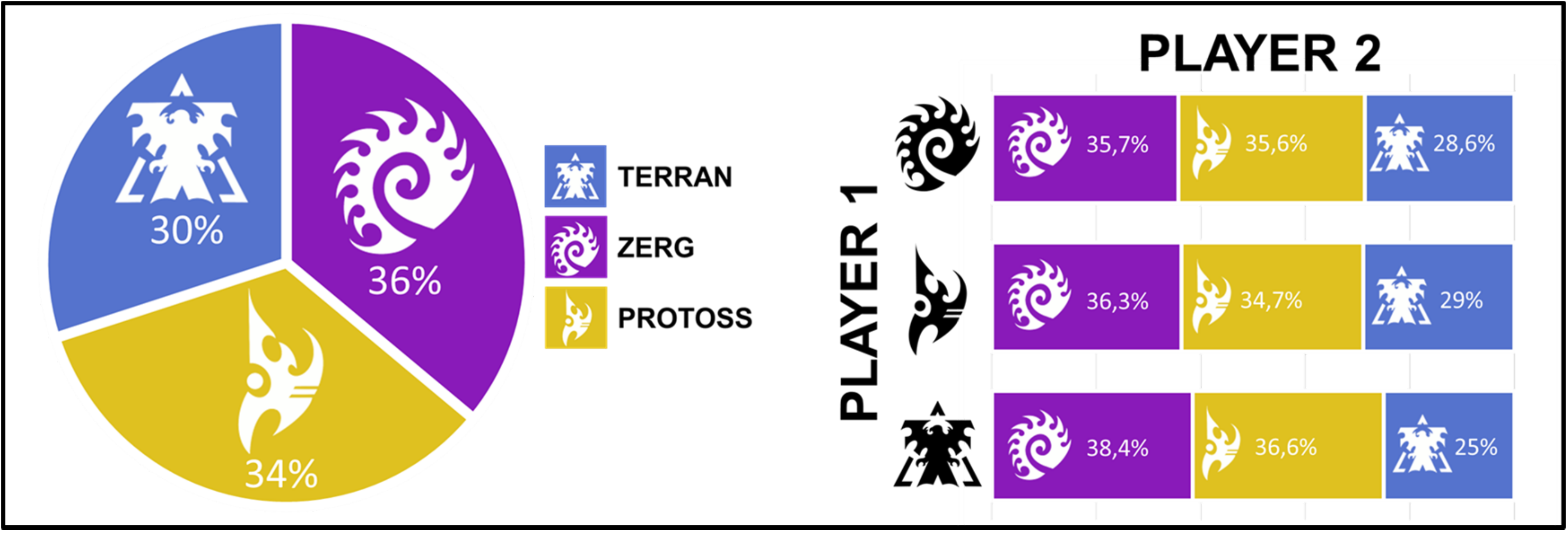}
    \caption{Distribution of player races and race matchup information.}
    \label{fig:SC2PlayerDistribution}
\end{figure}

\begin{wrapfigure}{R}{0pt}
    \centering
    \includegraphics[width=0.5\textwidth]{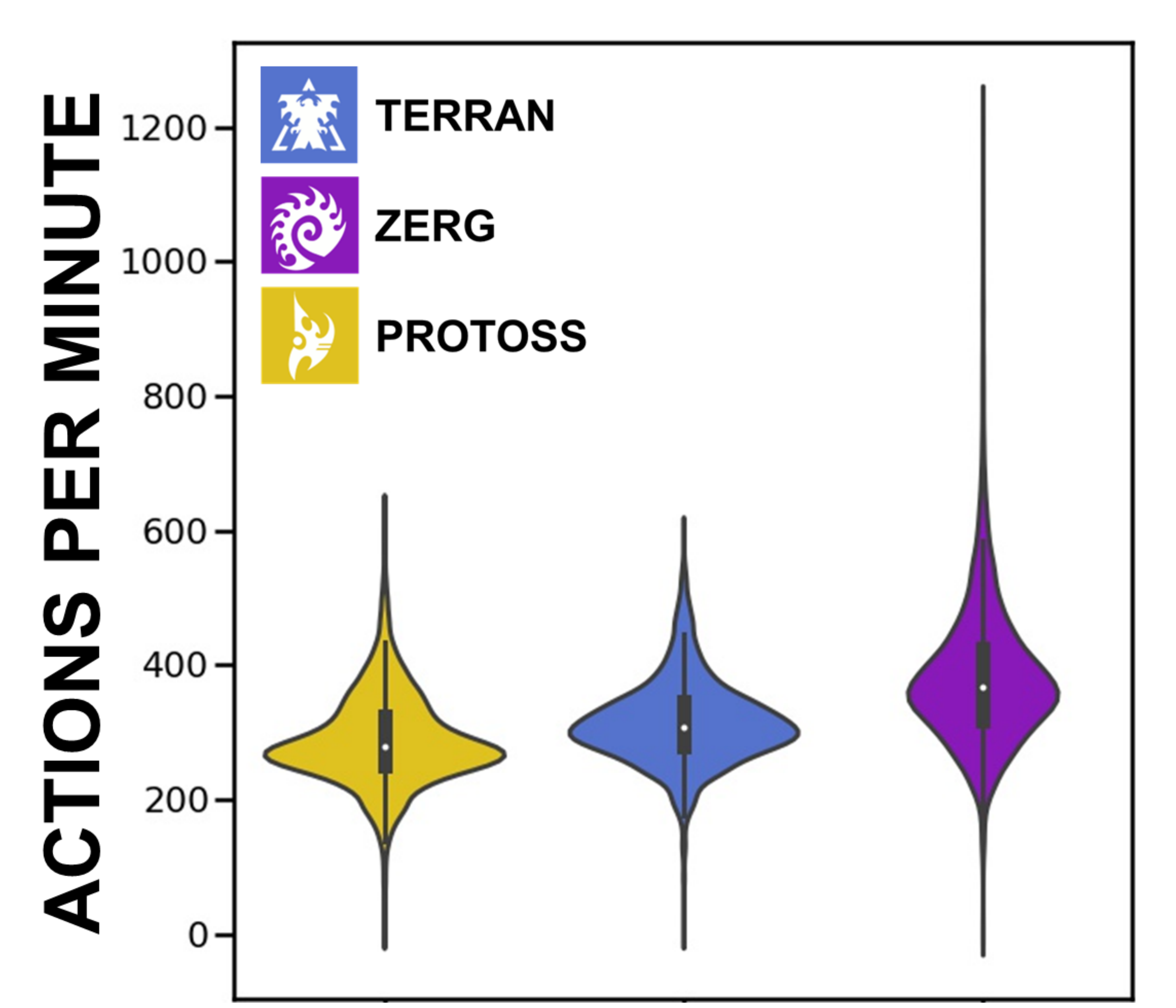}
    \caption{Actions per minute (APM)\\by player race.}
    \label{fig:SC2TimeHistogram}
\end{wrapfigure}

The collected dataset consisted of 55 tournaments. Within the available tournaments, 18309 matches were processed. The final processing yielded 17895 files. While inspecting the processed data, we observed three major 
game versions. Each tournament in the dataset was saved with an accompanying JSON file that contains descriptive statistics such as:
\begin{enumerate*}[label=(\arabic*)]
    \item Game version histogram,
    \item Dates at which the observed matches were played,
    \item Server information,
    \item Picked race information,
    \item Match length,
    \item Detected spawned units,
    \item Race picked versus game time histogram.
\end{enumerate*}
\autoref{fig:SC2PlayerDistribution} depicts the frequency with which each of the races played against the other and the distribution of races observed within the tournaments. \autoref{fig:SC2TimeHistogram} depicts the 
distribution of match times that were observed.

\begin{figure}[H]
    \centering
    \includegraphics[width=0.8\linewidth]{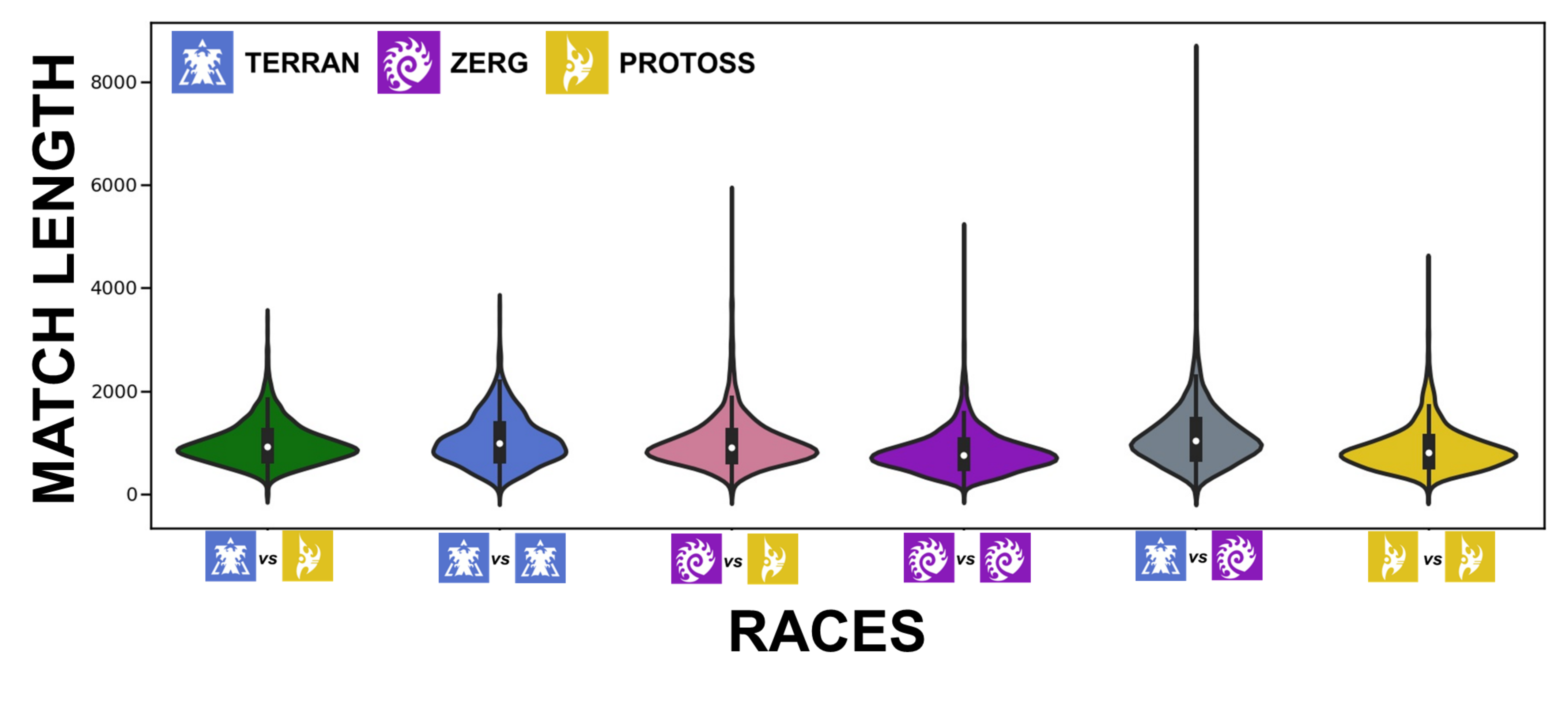}
    \caption{Match time distribution split by races: Terran (blue), Protoss (yellow), and Zerg (purple).}    
    \label{fig:SC2RaceTimeHistogram}
\end{figure}

The oldest observed tournament was IEM 10 Taipei, which was played in 2016. The most recent observed tournament was IEM Katowice, which finished on 2022.02.27. The game contains different "races" that differ in the mechanics required for the gameplay. \autoref{fig:SC2RaceTimeHistogram} shows visible differences in the distribution of match time for players that picked different races.

The published data resulting from our work is distributed under the Creative Commons Attribution-NonCommercial 4.0 International (CC BY-NC 4.0) license and is available in a widely recognized scientific repository - Zenodo.

\section{Experiments and Future Research}
\label{sec:Tasks_Experiments}

\subsection{Winner prediction and Player Performance Evaluation}
\label{sec:WinnerPredictionExperiment}

Within \autoref{sec:RelatedWork} we have referenced multiple articles that dealt with player performance evaluation. These works performed data mining tasks on game engine generated replays and other sources of player 
related information.

Experiments regarding winner prediction can uncover interesting information about the optimal strategy of play. Prior analyses in this task with a small sample of esports players have shown the importance of some key indicators. The proposed dataset can help with the reproduction and facilitation of various claims, some of which are based on anecdotal evidence \cite{Bialecki2021Determinants}. The sample analysis below describes a basic attempt at predicting match outcome using only data related to player economy to demonstrate the potential for gleaning insights from replay data.

\paragraph{Data Preparation}
Matches were initially filtered to only include those exceeding or equaling a length of 9 minutes, which is approximately the 25th percentile of match length values. Next, a set of features was generated from the available economy-related indicators. Additional features were generated by combining mineral and vespene indicators into summed resource indicators. Match data were then aggregated by averaging across match time for each player, resulting in 22,230 samples of averaged match data (from 11,115 unique matches). Standard deviations were computed in addition to averaged values where applicable. Further features were then generated by computing ratios of resources killed/resources lost for army, economy and technology contexts, along with a ratio of food made to food used. As a final step, prior to feature standardization, each feature was filtered for outliers (replacing with median) that exceeded an upper limit of 3 standard deviations from the feature mean.   

\paragraph{Feature Selection}
The feature count was reduced by first computing point biserial correlations between features and match outcome, selecting for features with a statistically significant (\(\alpha\) = .001) coefficient value exceeding that of \(\pm\) .050. Next, a matrix of correlations was computed for the remaining features and redundant features were removed. 17 features remained after this process, of which 8 were basic features (mineralsLostArmy, mineralsKilledArmy, mineralsLostEconomy, mineralsKilledEconomy, and the SD for each).

\paragraph{Modelling}
Modelling was conducted on features (economic indicators) that represented the global average gamestate, in which all time points were aggregated into a single state, and also as a time series in which the gamestate was represented at a sampling rate of approx. 7 seconds. Three algorithms were chosen for comparative purposes: Logistic Regression (sklearn.linear\_model.LogisticRegression), Support Vector Machine (sklearn.svm.SVC) \cite{scikit-learn,sklearnAPI}, and Extreme Gradient Boosting (xgboost.XGBClassifier) \cite{Chen2016}. Each algorithm was initiated with settings aimed at binary classification and with typical starting hyperparameters. A 5-fold cross validation procedure was implemented across the models.

Two sets of models were trained for the average gamestate and one for the gamestate as a time series. In the 
first averaged set of models the input features represented the economic gamestate of a single player without reference to their opponent, with the model output representing outcome prediction accuracy for that player 
- a binary classification problem on scalar win/loss classes. The second averaged set of models differed in that it used the averaged economic gamestate of both players as input features, and attempted to predict the outcome of "Player 1" for each match. Finally, the time series models consisted of a feature input vector containing the economic gamestate at 7 second intervals -- the task here was also to predict the outcome of a match based on only a single player's economic features, as in the single-player averaged set of models.        

Label counts were equalized to the minimal label count prior to generating the data folds, resulting in 10,744 samples of ``Win'' and ``Loss'' labels each for the single-player averaged models and the time series models. For the two-player set of averaged models (containing the features of both players in a given match), the 
total number of matches used was 10,440. Accuracy was chosen as the model performance evaluation metric in all three cases. Computation was performed on a standard desktop-class PC without additional resources.        

\paragraph{Results}
As the results indicate (see \autoref{table:ClassificationPerformance}), good outcome prediction can be achieved from economic indicators only, even without exhaustive optimization of each model's hyperparameters. For 
the one-player averaged set of models, SVM and XGBoost displayed similar performance, with the logistic classifier lagging slightly behind. For the two-player averaged set of models, all three algorithms performed essentially equally well. Feature importances were taken from a single-player XGBoost model (with identical hyperparameters) that was applied to the entire dataset for illustrative purposes. \autoref{fig:SC2FeatureImportance} below depicts the top five features by importance. It is interesting to note that importance was more heavily centered around mineral-related features than those tied to vespene, which is likely tied to how mineral and vespene needs are distributed across unit/building/technology costs. Further feature investigation is required to verify this tendency.

\begin{table}[H]
    \centering
    \caption{Classification models and their performance metrics for two separate win prediction models. The 
``One Player Prediction'' models attempt to correctly output if one of the players won or lost. The ``Two Player Prediction'' models have access to the data for both of the players and attempts to output if "Player 1" 
won or lost.}
    \label{table:ClassificationPerformance}
    \resizebox{0.91\columnwidth}{!}{
        \begin{tabular}{llll}
            \hline
            Classifier                   & \multicolumn{1}{c}{Accuracy} & \multicolumn{1}{c}{SD} & \multicolumn{1}{c}{Hyperparameters}     \\ \hline
            \multicolumn{4}{c}{One Player Prediction}
                              \\ \hline
            Support Vector Machine - RBF & 0.8488                       & 0.0075                 & kernel='rbf', C=10, gamma='auto'        \\
            XGBoost                      & 0.8397                       & 0.0064                 & Booster='gbtree', eta=0.2, max\_depth=5 \\
            Logistic Regression          & 0.8118                       & 0.0057                 & C=10, penalty='l2'                      \\ \hline
            \multicolumn{4}{c}{Two Player Prediction}
                              \\ \hline
            Support Vector Machine - RBF & 0.9071                       & 0.0055                 & kernel='rbf', C=10, gamma='auto'        \\
            XGBoost                      & 0.8924                       & 0.0063                 & Booster='gbtree', eta=0.2, max\_depth=5 \\
            Logistic Regression          & 0.8916                       & 0.0063                 & C=10, penalty='l2'                      \\ \hline
            \end{tabular}
    }
\end{table}

\begin{figure}[H]
    \centering
    \includegraphics[width=0.7\linewidth]{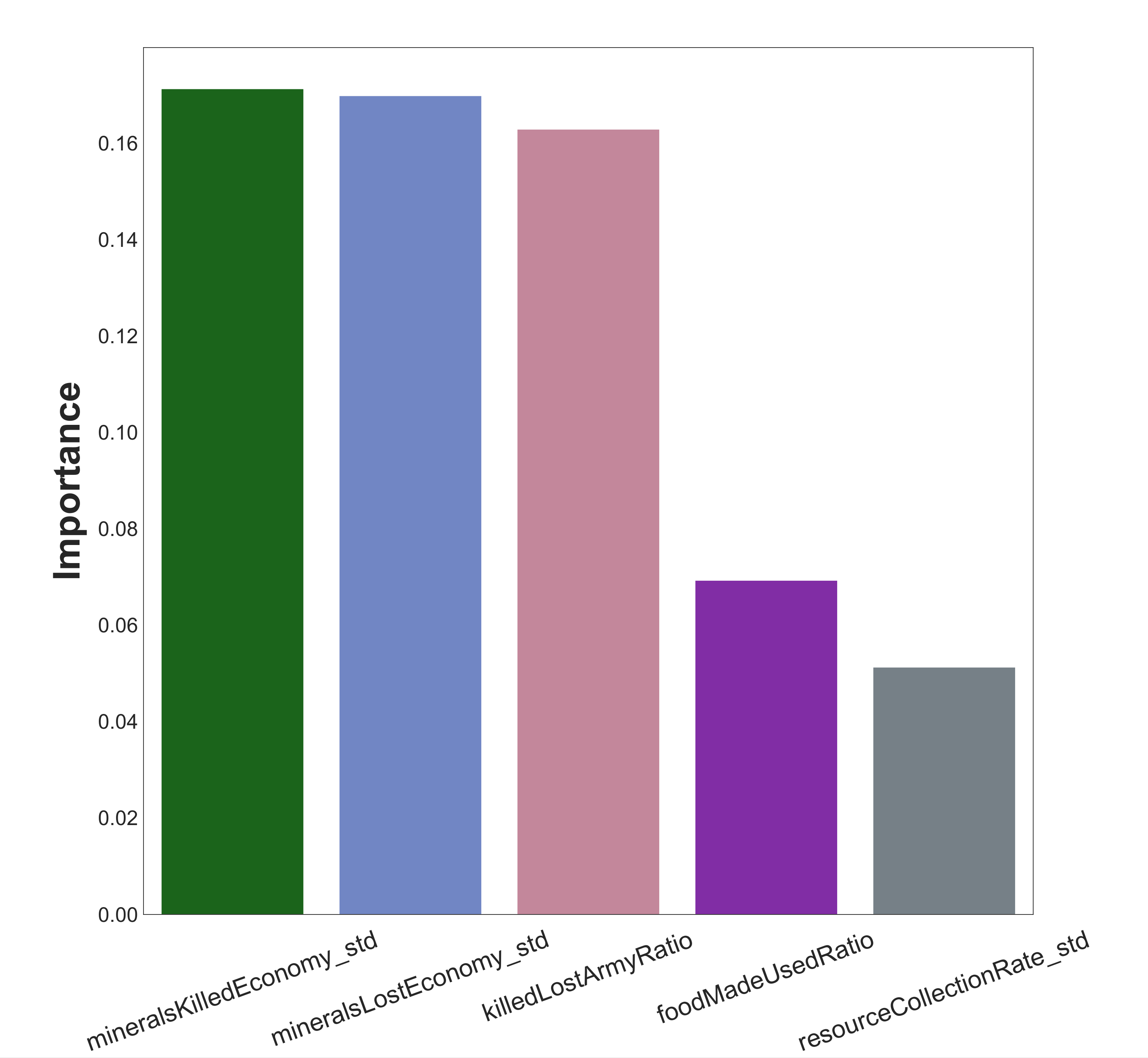}
    \caption{Percentages of feature importances based on XGBoost fit to all data.}
    \label{fig:SC2FeatureImportance}
\end{figure}

\begin{figure}[H]
    \centering
    \includegraphics[width=0.75\linewidth]{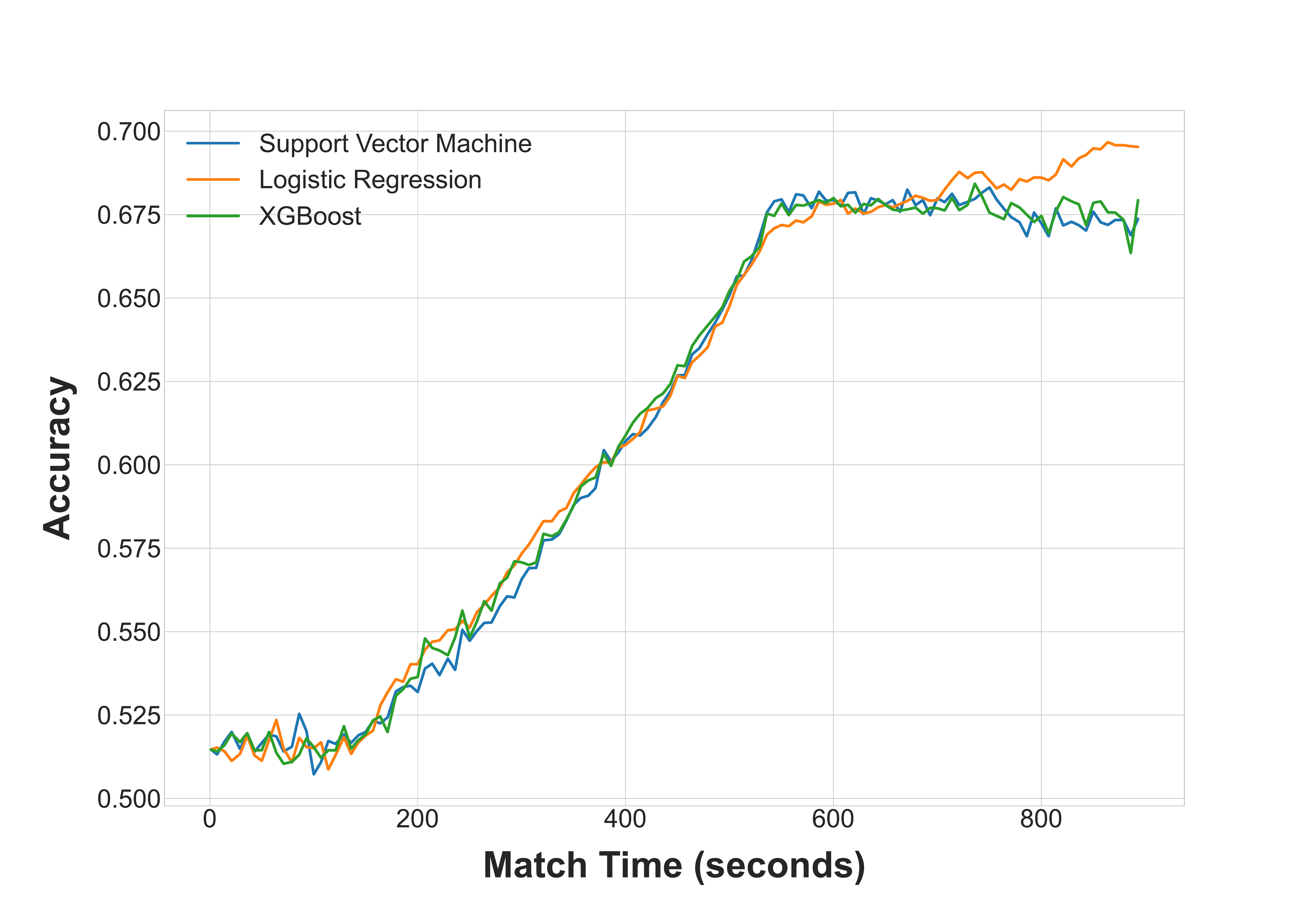}
    \caption{Accuracy comparison of applied classification models.}
    \label{fig:SC2ClassifierComparison}
\end{figure}

\autoref{fig:SC2ClassifierComparison} depicts the time series application of these models as an illustration 
of outcome prediction accuracy over time. It should be noted that these time series results are not based on 
any form of data aggregation, and as such only basic economic features could be used for classification (18 features in total).

Each timepoint contains the average accuracy for 5-fold cross validation, with a minimum match length of 9 minutes and a maximum match length of approx. 15 minutes. All three algorithms provided similar performance over time, although this may be an effect of the minimal hyperparameter optimization that was performed. Further, it is also interesting to note and that all three algorithms meet a classification performance asymptote at approx. the same match time (\textasciitilde550 seconds), which may indicate that this is where economic indicators begin to lose their predictive power and (presumably) other factors such as army size, composition, and their application become the primary determinants. The code for our experiments is available at a dedicated GitHub repository \cite{PawelExperiments}.

\subsection{Future Research}

\subsubsection{Game Style Analysis}

Game style analysis can be treated as a task to be solved via supervised or self-supervised methods. Using algorithms such as Uniform Manifold Approximation and Projection (UMAP) \cite{McInnes2018} or t-Distributed Stochastic Neighbor Embedding (t-SNE) \cite{Laurens2008} for the data that we provided could uncover interesting insights depending on the direction of the analysis. Such game style analysis could be investigated using sequence analysis methods or use per game statistics.

\subsubsection{Combat Encounter Analysis}

Combat analysis as a task can be researched using AI, ML, and classic algorithms in various esports \cite{Uriarte2018}. There were some related works that analyzed unit encounters in StarCraft II \cite{Lee2021}. Although our pre-processed dataset cannot be used to directly reproduce combat encounter analyses, we provide raw replays published as SC2ReSet \cite{BialeckiSC2ReSet2021}.

\section{Limitations}
\label{sec:Limitations}

We acknowledge that our work is not without limitations. The design and implementation of our dataset do not 
consider the ability to obtain StarCraft II data through game-engine simulation at a much higher resolution. 
Because of this, the extracted dataset cannot reflect exact unit positioning.
Replays in their original MPQ (SC2Replay) format contain all necessary information to recreate a game using game-engine API. Therefore, we plan to continue our research and provide more datasets that will expand the scientific possibilities within gaming and esports. Further, it should be noted that the experiments described 
here are more illustrative than investigative in nature, and could be greatly expanded upon in future work.  
We recommend further research to use SC2ReSet \cite{BialeckiSC2ReSet2021} to compute game-engine simulated information.
We also do not provide simulation observation data that allows more detailed spatiotemporal information to be extracted at a higher computational cost. Moreover, it is worth noting that the dataset completeness was dependent on which tournament organizers and tournament administrators decided to publish replay packs.

\section{Discussion}
\label{sec:Discussion}

Future authors may want to filter out replays that ended prematurely due to unknown reasons. Our dataset may 
contain replays that are undesirable for esports research. We have decided against the deletion of replays to preserve the initial distributions of data. Additionally, as filtering was omitted (besides that performed for the purposes of the described experiments), there is a risk that the dataset contains matches that were a 
part of the tournament itself but did not count towards the tournament standings. Due to the timeframe of the tournaments and game version changes, despite our best efforts, some information might be missing or corrupted and is subject to further processing and research.

Our dataset is the largest publicly available pre-processed esports dataset. Moreover, in preparing the data, we defined and published the software used for the data extraction process and other tasks. Future research 
on StarCraft II may be built upon these tools and our dataset \cite{Bialecki2021Preparator,Bialecki2021MapExtractor,Bialecki2021InfoExtractor}.

The dataset may also serve to increase knowledge regarding the in-game behavior of players, i.e. the relationship between the variables and overall strategies used by the players at high levels of advancement. Such information can be used in comparisons to non-gamers or intermediate players in the process of studying the relationship between game proficiency, cognitive functioning, and brain structure \cite{Jakubowska2021}.

Moreover, a report done in the area of clinical medicine highlighted the lack of compliance of many authors with their data availability statement (DAS). It is clear that publishing the data and tools required for modeling is a key component of ensuring reproducible scientific work \cite{Gabelica2022}.

Other noteworthy applications of the dataset include comparing gameplay styles, action sequence classification, and their relation to victory. To that end, we encourage using different statistical methods and Machine Learning (ML) algorithms, including supervised and self-supervised approaches.

\section*{Acknowledgements}

We would like to acknowledge various contributions by the members of the technical and research community, with special thanks to: Timo Ewalds (DeepMind, Google), Anthony Martin (Sc2ReplayStats), and András Belicza for assisting with our technical questions.
Moreover, we extend our thanks to the StarCraft II esports community for sharing their experiences, playing together, and discussing key aspects of the gameplay in various esports. We extend our thanks especially to: Mikołaj ``Elazer'' Ogonowski, Konrad ``Angry'' Pieszak, Mateusz ``Gerald'' Budziak, Igor ``Indy'' Kaczmarek, Adam ``Ziomek'' Skorynko, Jakub ``Trifax'' Kosior, Michał ``PAPI'' Królikowski, and Damian ``Damrud'' Rudnicki.

\section*{Declarations}

\subsection*{Authors' contributions}

\begin{itemize}
    \item Conceptualization: Andrzej Białecki;
    \item Supervision: Andrzej Białecki, Jan Gajewski;
    \item Methodology: Andrzej Białecki, Natalia Jakubowska, Paweł Dobrowolski, Piotr Białecki, Leszek Krupiński;
    \item Formal Analysis: Andrzej Białecki, Natalia Jakubowska, Paweł Dobrowolski;
    \item Investigation: Andrzej Białecki, Natalia Jakubowska, Paweł Dobrowolski, Piotr Białecki, Robert Białecki;
    \item Writing - original draft: Andrzej Białecki;
    \item Writing - review and editing: Andrzej Białecki, Paweł Dobrowolski, Andrzej Szczap, Robert Białecki,\\ Jan Gajewski;
    \item Data curation: Andrzej Białecki, Andrzej Szczap;
    \item Technical Oversight: Piotr Białecki;
    \item Software: Andrzej Białecki, Leszek Krupiński;
    \item Technical Documentation: Andrzej Szczap
\end{itemize}

\subsection*{Author statement}
We acknowledge that we as authors bear all responsibility in case of violation of rights.

\subsection*{Funding}
This publication was self-funded.
\subsection*{Conflicts of interest/Competing interests}
Authors declare no conflict of interest.
\subsection*{Availability of data and material}
Extracted data is published as a dataset in a scientific repository \cite{BialeckiEGSetDataset,BialeckiSC2ReSet2021}.
\subsection*{Code Availability}
The code used for data extraction is available as open source implementations published by the authors \cite{Bialecki2021Preparator,Bialecki2021MapExtractor,Bialecki2021InfoExtractor}. The code used for experiments is available for preview 
in a GitHub repository \cite{PawelExperiments}.

In the process of preparing this article, PyTorch Lightning has changed its name into Lightning. We have decided to use 
the old form of the name, following the citation template provided by the Lightning project on GitHub \cite{PyTorch_Lightning_2019}.

\bibliographystyle{IEEEtran}
\bibliography{sc2egset_pre_print_collected.bib}

\appendix

\section{Appendix}
\label{sec:appendix}

\subsection{Blizzard StarCraft II AI and Machine Learning License}
\label{sec:AILicense}

BLIZZARD® STARCRAFT® II AI AND MACHINE LEARNING LICENSE

IMPORTANT NOTICE:

YOU SHOULD CAREFULLY READ THIS AGREEMENT (THE “AGREEMENT”) BEFORE INSTALLING OR USING
BLIZZARD’S (“BLIZZARD”) STARCRAFT II AI AND MACHINE LEARNING SOFTWARE AND ENVIRONMENT (THE
“SOFTWARE”).  IF YOU DO NOT AGREE WITH ALL OF THE TERMS OF THIS AGREEMENT, YOU MAY NOT INSTALL
OR OTHERWISE ACCESS THE SOFTWARE.

Subject to the terms of this Agreement, your use of the Software is governed by Blizzard’s End
User License Agreement (“EULA”), which is incorporated by reference herein and is available
for review here. (\href{http://us.blizzard.com/en-us/company/legal/eula.html}{http://us.blizzard.com/en-us/company/legal/eula.html})  Please carefully
review the EULA and this Agreement prior to installing or using the Software.  IF YOU DO NOT
AGREE TO THE TERMS OF THE EULA AND THIS AGREEMENT, YOU ARE NOT PERMITTED TO INSTALL, COPY, OR
USE THE SOFTWARE.

\begin{enumerate}[label=\arabic*.]
    \item Use Of The Software.
    \begin{enumerate}[label=\Alph*.]
        \item AI Testing And Machine Learning Use Only: Subject to your compliance with this
        Agreement, Blizzard grants you a limited, revocable, non-sublicensable license to use the
        Software for purposes of AI testing, machine learning, and related research only.
        \item Blizzard Account Not Required: Notwithstanding the requirements of Section 1.A of the
        EULA, creation of a Blizzard Account is not required in order to use the Software.  Legal
        entities other than natural persons are authorized to use the Software.  However, other
        than as specifically excepted in this Agreement, the remaining provisions and requirements
        of the EULA are controlling.
        \item EULA Exceptions: The terms of Blizzard's EULA govern your use of the Software, subject
        to the following narrow exceptions:
        \begin{enumerate}[label=\roman*.]
            \item Derivative Works: Section 1.C.i of the EULA shall not be read to prohibit the
            authorized use of the Software or data generated or collected from such use. However,
            no portion of this Agreement shall give you the right to create, distribute, or
            otherwise exploit unauthorized derivative works of the Software.
            \item Automation: The provisions of Section 1.C.ii of the EULA prohibiting the use of
            automation processes or software do not apply to use of the Software.
            \item Commercial Use:  The provisions of Section 1.C.iii of the EULA govern your use
            of the Software, except that you are authorized to use and exploit data derived from
            using the Software in connection with AI and machine learning programs for personal or
            internal use, despite that such use of the data may ultimately be for a commercial
            purpose.  You may not otherwise use or exploit the Software for any commercial purpose.
            \item Data Mining: The provisions of Section 1.C.iv of the EULA shall not prohibit the
            authorized use of the Software or data generated or collected from such use.
            \item Matchmaking: The provisions of Section 1.C.vi of the EULA shall not prohibit the
            authorized use of the Software or data generated or collected from such use.
        \end{enumerate}
    \end{enumerate}
    \item Ownership.
    \begin{enumerate}[label=\Alph*.]
        \item The provisions of Section 2 of the EULA apply in full force to the Software (including
        generated by or collected through the authorized use of the Software.
    \end{enumerate}
\end{enumerate}

\subsection{Datasheets for datasets}

Datasheets for datasets \cite{Gebru2018} are defined and available as a part of the original pre-processed dataset publication. \cite{BialeckiEGSetDataset}

\subsection{Pipeline Visualizations}
\label{sec:appPipelineVisualizations}

\begin{figure}[H]
    \includegraphics[width=\linewidth]{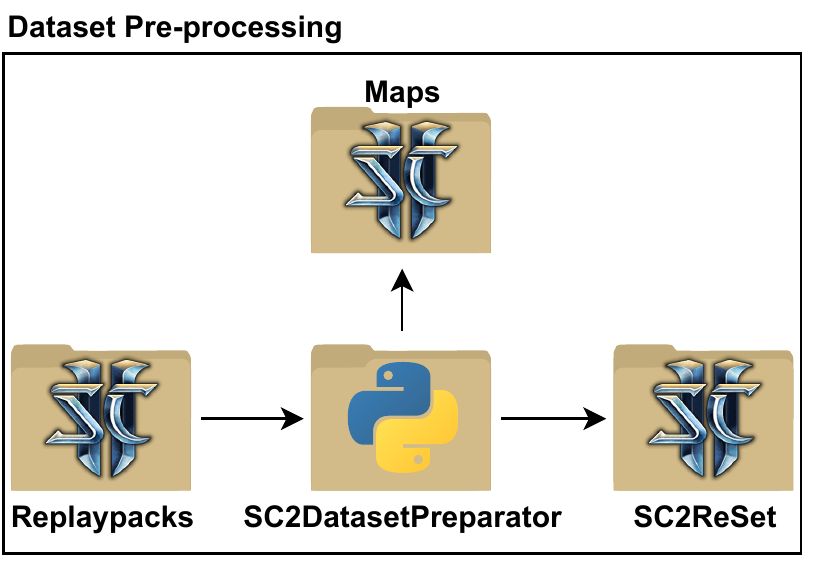}
    \caption{Pre-processing steps of our pipeline that result in SC2ReSet \cite{BialeckiSC2ReSet2021}. We are using a custom data processing toolset including SC2DatasetPreparator \cite{Bialecki2021Preparator}.}
    \label{fig:DatasetPreProcess}
\end{figure}

\begin{figure}[H]
    \includegraphics[width=\linewidth]{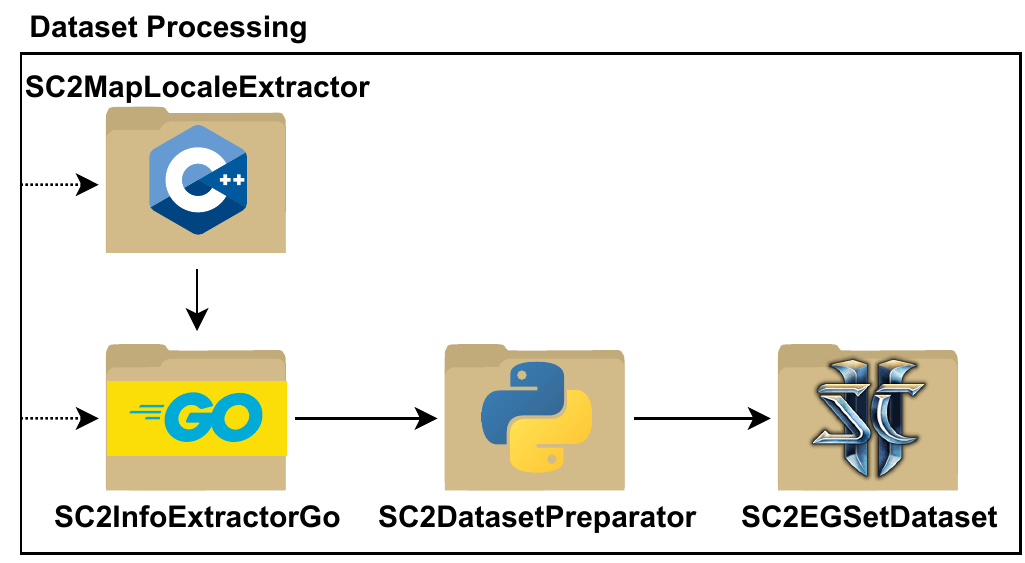}
    \caption{Processing steps of our pipeline that result in SC2EGSetDataset \cite{BialeckiEGSetDataset}. We are using a custom data processing toolset including SC2DatasetPreparator \cite{Bialecki2021Preparator}, SC2MapLocaleExtractor \cite{Bialecki2021MapExtractor}, and SC2InfoExtractorGo \cite{Bialecki2021InfoExtractor}.}
    \label{fig:DatasetProcess}
\end{figure}

\begin{figure}[H]
    \includegraphics[width=\linewidth]{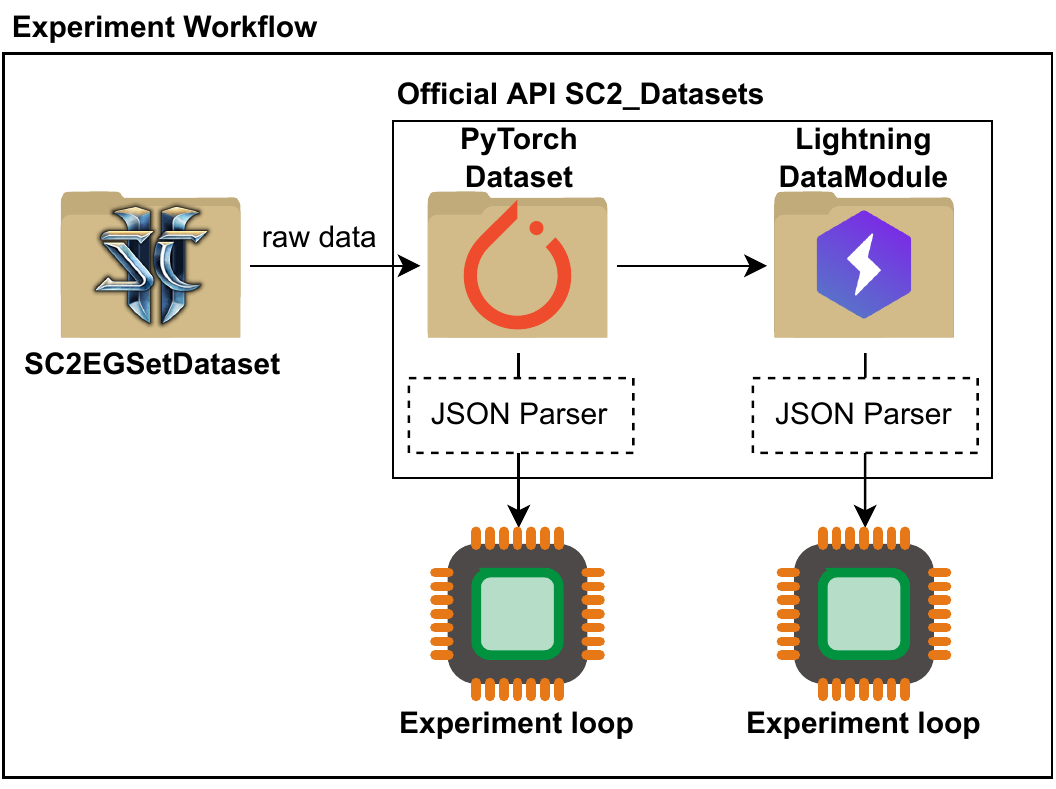}
    \caption{Using the SC2EGSetDataset \cite{BialeckiEGSetDataset} with the officially provided API \cite{BialeckiDatasetAPI} to conduct experiments.}
    \label{fig:DatasetPostProcess}
\end{figure}

Due to the relative complexity of our infrastructure, we include additional visualizations of the processing pipeline for all potential users on Figures \ref{fig:DatasetPreProcess} - \ref{fig:DatasetPostProcess}. The dataset pre-processing 
is shown on \autoref{fig:DatasetPreProcess}, highlighting the use of a set of tools named SC2DatasetPreparator \cite{Bialecki2021Preparator}. The dataset processing is shown on \autoref{fig:DatasetProcess}, highlighting the use of SC2MapLocaleExtractor \cite{Bialecki2021MapExtractor} to acquire the english map names, SC2InfoExtractorGo \cite{Bialecki2021InfoExtractor} to extract the data, and SC2DatasetPreparator \cite{Bialecki2021Preparator} to collect the final .zip archives. The dataset post-processing and experiments are briefly visualized on \autoref{fig:DatasetPostProcess} and highlight 
the use of PyTorch \cite{PyTorch2019}, and Lightning \cite{PyTorch_Lightning_2019}.

\subsection{Dataset Usage Examples}
\label{sec:appDatasetUsage}

There are various ways to use our dataset; one way includes using the custom PyTorch dataset class which was briefly introduced in \autoref{sec:DatasetLoading}, \autoref{lst:ExamplePyTorch}. Due to the page limit we were unable to visualize all of the potential uses of our infrastructure in the main text; \autoref{lst:ExampleLightning} showcases the most basic use of the Lightning custom DataModule class that we implemented for our dataset. For further information please refer to the official documentation.

\begin{adjustbox}{max width=\textwidth}
    \begin{lstlisting}[caption={Example use of the SC2EGSetDataModule with Lightning using a synthetic replaypack prepared for testing.},label={lst:ExampleLightning},language=python,style=mypython,firstnumber=1]
from sc2_datasets.lightning.sc2_egset_datamodule import (
    SC2EGSetDataModule
)
from sc2_datasets.available_replaypacks import (
    EXAMPLE_SYNTHETIC_REPLAYPACKS
)

if __name__ == "__main__":
    # Initialize the datamodule:
    sc2_egset_datamodule = SC2EGSetDataModule(
        unpack_dir="./unpack_dir_path",
        download_dir="./download_dir_path",
        download=True,
        replaypacks=EXAMPLE_SYNTHETIC_REPLAYPACKS,
    )

    # Initializing the PyTorch dataset within the DataModule class:
    sc2_egset_datamodule.prepare_data()
    # Obtaining the splits for training, testing, and validation:
    sc2_egset_datamodule.setup()
\end{lstlisting}
\end{adjustbox}

It is important to note that our classes by default return a custom SC2ReplayData class which is a serialization of the 
initial pre-processed JSON files. To construct a custom tensor required for further modeling, users should use the exposed keyword argument ``transform'', which should be a function that transforms the default SC2ReplayData into some custom tensor required for further modeling.

\subsection{Dataset Structure Examples}
\label{sec:appDatasetStructureExamples}

We include human-readable examples of various fields showcase on Listings \ref{lst:JSONHeaderListing}-\ref{lst:UnitPositionsListing}; these are a part of the SC2EGSet dataset JSON files. Users should refer to the respective parts of the official documentation for more information and a list of all of the available fields. Access to these can be used to define parsers in other programming languages.

\subsubsection{Top level fields}

\begin{minipage}{\linewidth}
\begin{lstlisting}[caption={Example header field containing a JSON object.},label={lst:JSONHeaderListing},language=json,firstnumber=1]
{
    ...
    "header": {
        "elapsedGameLoops": 7855,
        "version": "3.4.0.44401"
    },
    ...
}
\end{lstlisting}
\end{minipage}

\begin{minipage}{\linewidth}
\begin{lstlisting}[caption={Example initData field containing a JSON object with nested information.},language=json,firstnumber=1]
{
    ...
    "initData": {
        "gameDescription": {
            "gameOptions": {
                "advancedSharedControl": false,
                "amm": false,
                "battleNet": true,
                "clientDebugFlags": 265,
                "competitive": false,
                "cooperative": false,
                "fog": 0,
                "heroDuplicatesAllowed": true,
                "lockTeams": true,
                "noVictoryOrDefeat": false,
                "observers": 0,
                "practice": false,
                "randomRaces": false,
                "teamsTogether": false,
                "userDifficulty": 0
            },
            "gameSpeed": "Faster",
            "isBlizzardMap": true,
            "mapAuthorName": "5-S2-1-1",
            "mapFileSyncChecksum": 360400735,
            "mapSizeX": 152,
            "mapSizeY": 152,
            "maxPlayers": 2
        }
    },
    ...
}
\end{lstlisting}
\end{minipage}

\begin{minipage}{\linewidth}
\begin{lstlisting}[caption={Example details field containing a JSON object.},language=json,firstnumber=1]
{
    ...
    "details": {
        "gameSpeed": "Faster",
        "isBlizzardMap": true,
        "timeUTC": "2016-07-29T04:50:12.5655603Z"
    },
    ...
}
\end{lstlisting}
\end{minipage}

\begin{minipage}{\linewidth}
\begin{lstlisting}[caption={Example metadata field containing a JSON object.},language=json,firstnumber=1]
{
    ...
    "metadata": {
        "baseBuild": "",
        "dataBuild": "",
        "gameVersion": "",
        "mapName": "Galactic Process LE"
    },
    ...
}
\end{lstlisting}
\end{minipage}

\begin{minipage}{\linewidth}
\begin{lstlisting}[caption={Example metadata field containing a JSON object.},language=json,firstnumber=1]
    {
        ...
        "metadata": {
            "baseBuild": "",
            "dataBuild": "",
            "gameVersion": "",
            "mapName": "Galactic Process LE"
        },
        ...
    }
\end{lstlisting}
\end{minipage}

\begin{minipage}{\linewidth}
\begin{lstlisting}[caption={Example ToonPlayerDescMap field containing a JSON object mapping player statistics to unique toon id.},language=json,firstnumber=1]
{
    ...
    "ToonPlayerDescMap": {
        "5-S2-1-7361539": {
            "nickname": "somePlayerNickname",
            "playerID": 2,
            "userID": 5,
            "SQ": 105,
            "supplyCappedPercent": 4,
            "startDir": 1,
            "startLocX": 127,
            "startLocY": 131,
            "race": "Zerg",
            "selectedRace": "",
            "APM": 0,
            "MMR": 0,
            "result": "Win",
            "region": "China",
            "realm": "China",
            "highestLeague": "Unknown",
            "isInClan": false,
            "clanTag": "",
            "handicap": 100,
            "color": {
            "a": 255,
            "b": 0,
            "g": 66,
            "r": 255
            }
        },
        "5-S2-1-7361634": {
            "nickname": "AnotherPlayerNickname",
            "playerID": 1,
            "userID": 1,
            "SQ": 115,
            "supplyCappedPercent": 7,
            "startDir": 7,
            "startLocX": 24,
            "startLocY": 20,
            "race": "Zerg",
            "selectedRace": "",
            "APM": 0,
            "MMR": 0,
            "result": "Loss",
            "region": "China",
            "realm": "China",
            "highestLeague": "Unknown",
            "isInClan": false,
            "clanTag": "",
            "handicap": 100,
            "color": {
            "a": 255,
            "b": 180,
            "g": 20,
            "r": 30
            }
        }
    }
    ...
}
\end{lstlisting}
\end{minipage}

\subsubsection{Game Events}

All of the game events that were recorded by the game engine are available in one of the fields named \lstinline[language=json]!"gameEvents"!, all of the events that are available are presented in \dots

\begin{minipage}{\linewidth}
\begin{lstlisting}[caption={Example UserOptions game event JSON object.},language=json,firstnumber=1]
[
    ...
    {
        "baseBuildNum": 44401,
        "buildNum": 44401,
        "cameraFollow": false,
        "debugPauseEnabled": false,
        "developmentCheatsEnabled": false,
        "evtTypeName": "UserOptions",
        "gameFullyDownloaded": true,
        "hotkeyProfile": "\u003ccustom\u003e",
        "id": 7,
        "isMapToMapTransition": false,
        "loop": 0,
        "multiplayerCheatsEnabled": false,
        "platformMac": false,
        "syncChecksummingEnabled": false,
        "testCheatsEnabled": false,
        "useGalaxyAsserts": false,
        "userid": {
                "userId": 0
            },
        "versionFlags": 0
        },
    ...
]
\end{lstlisting}
\end{minipage}

\begin{minipage}{\linewidth}
\begin{lstlisting}[caption={Example CameraUpdate game event JSON object.},language=json,firstnumber=1]
[
    ...
    {
        "distance": null,
        "evtTypeName": "CameraUpdate",
        "follow": false,
        "id": 49,
        "loop": 2,
        "pitch": null,
        "reason": null,
        "target": {
            "x": 0.7109375,
            "y": 0.5469970703125
        },
        "userid": {
            "userId": 6
        },
        "yaw": null
    },
    ...
]
\end{lstlisting}
\end{minipage}

\begin{minipage}{\linewidth}
\begin{lstlisting}[caption={Example SelectionDelta game event JSON object.},language=json,firstnumber=1]
[
    ...
    {
        "controlGroupId": 10,
        "delta": {
            "addSubgroups": [
            {
                "count": 1,
                "intraSubgroupPriority": 1,
                "subgroupPriority": 32,
                "unitLink": 108
            }
            ],
            "addUnitTags": [
            56885249
            ],
            "removeMask": {
            "None": null
            },
            "subgroupIndex": 0
        },
        "evtTypeName": "SelectionDelta",
        "id": 28,
        "loop": 12,
        "userid": {
            "userId": 5
        }
        },
    ...
]
\end{lstlisting}
\end{minipage}

\begin{minipage}{\linewidth}
\begin{lstlisting}[caption={Example Cmd game event JSON object.},language=json,firstnumber=1]
[
    ...
    {
        "abil": {
            "abilCmdData": null,
            "abilCmdIndex": 0,
            "abilLink": 188
        },
        "cmdFlags": 256,
        "data": {
            "None": null
        },
        "evtTypeName": "Cmd",
        "id": 27,
        "loop": 15,
        "otherUnit": null,
        "sequence": 1,
        "unitGroup": null,
        "userid": {
            "userId": 5
        }
    },
    ...
]
\end{lstlisting}
\end{minipage}

\begin{minipage}{\linewidth}
\begin{lstlisting}[caption={Example CmdUpdateTargetUnit game event JSON object.},language=json,firstnumber=1]
[
    ...
    {
        "evtTypeName": "CmdUpdateTargetUnit",
        "id": 105,
        "loop": 37,
        "target": {
            "snapshotControlPlayerId": 0,
            "snapshotPoint": {
            "x": 64.5,
            "y": 68.75,
            "z": 5.994140625
            },
            "snapshotUnitLink": 369,
            "snapshotUpkeepPlayerId": 0,
            "tag": 2883585,
            "targetUnitFlags": 111,
            "timer": 0
        },
        "userid": {
            "userId": 5
        }
    },
    ...
]
\end{lstlisting}
\end{minipage}

\begin{minipage}{\linewidth}
\begin{lstlisting}[caption={Example CommandManagerState game event JSON object.},language=json,firstnumber=1]
[
    ...
    {
        "evtTypeName": "CommandManagerState",
        "id": 103,
        "loop": 37,
        "sequence": 3,
        "state": 1,
        "userid": {
            "userId": 5
        }
    },
    ...
]
\end{lstlisting}
\end{minipage}

\begin{minipage}{\linewidth}
\begin{lstlisting}[caption={Example ControlGroupUpdate game event JSON object.},language=json,firstnumber=1]
[
    ...
    {
        "controlGroupIndex": 1,
        "controlGroupUpdate": 2,
        "evtTypeName": "ControlGroupUpdate",
        "id": 29,
        "loop": 1639,
        "mask": {
            "None": null
        },
        "userid": {
            "userId": 1
        }
    },
    ...
]
\end{lstlisting}
\end{minipage}

\begin{minipage}{\linewidth}
\begin{lstlisting}[caption={Example CmdUpdateTargetPoint game event JSON object.},language=json,firstnumber=1]
[
    ...
    {
        "evtTypeName": "CmdUpdateTargetPoint",
        "id": 104,
        "loop": 2965,
        "target": {
            "x": 19.133056640625,
            "y": 26.369140625,
            "z": 5.73388671875
        },
        "userid": {
            "userId": 5
        }
    },
    ...
]
\end{lstlisting}
\end{minipage}

\begin{minipage}{\linewidth}
\begin{lstlisting}[caption={Example CmdUpdateTargetPoint game event JSON object.},language=json,firstnumber=1]
[
    ...
    {
        "evtTypeName": "GameUserLeave",
        "id": 101,
        "leaveReason": 0,
        "loop": 7845,
        "userid": {
            "userId": 5
        }
    },
    ...
]
\end{lstlisting}
\end{minipage}

\subsubsection{Tracker Events}

All of the game events that were recorded by the game engine are available in one of the fields named \lstinline[language=json]!"trackerEvents"!, all of the events that are available are presented in \dots

\begin{minipage}{\linewidth}
\begin{lstlisting}[caption={Example PlayerSetup tracker event JSON object.},language=json,firstnumber=1]
[
    ...
    {
        "evtTypeName": "PlayerSetup",
        "id": 9,
        "loop": 0,
        "playerId": 1,
        "slotId": 0,
        "type": 1,
        "userId": 1
    },
    ...
]
\end{lstlisting}
\end{minipage}

\begin{minipage}{\linewidth}
\begin{lstlisting}[caption={Example PlayerStats tracker event JSON object.},language=json,firstnumber=1]
[
    ...
    {
        "evtTypeName": "PlayerStats",
        "id": 0,
        "loop": 1,
        "playerId": 1,
        "stats": {
            "scoreValueFoodMade": 57344,
            "scoreValueFoodUsed": 49152,
            "scoreValueMineralsCollectionRate": 0,
            "scoreValueMineralsCurrent": 50,
            "scoreValueMineralsFriendlyFireArmy": 0,
            "scoreValueMineralsFriendlyFireEconomy": 0,
            "scoreValueMineralsFriendlyFireTechnology": 0,
            "scoreValueMineralsKilledArmy": 0,
            "scoreValueMineralsKilledEconomy": 0,
            "scoreValueMineralsKilledTechnology": 0,
            "scoreValueMineralsLostArmy": 0,
            "scoreValueMineralsLostEconomy": 0,
            "scoreValueMineralsLostTechnology": 0,
            "scoreValueMineralsUsedActiveForces": 0,
            "scoreValueMineralsUsedCurrentArmy": 0,
            "scoreValueMineralsUsedCurrentEconomy": 1050,
            "scoreValueMineralsUsedCurrentTechnology": 0,
            "scoreValueMineralsUsedInProgressArmy": 0,
            "scoreValueMineralsUsedInProgressEconomy": 0,
            "scoreValueMineralsUsedInProgressTechnology": 0,
            "scoreValueVespeneCollectionRate": 0,
            "scoreValueVespeneCurrent": 0,
            "scoreValueVespeneFriendlyFireArmy": 0,
            "scoreValueVespeneFriendlyFireEconomy": 0,
            "scoreValueVespeneFriendlyFireTechnology": 0,
            "scoreValueVespeneKilledArmy": 0,
            "scoreValueVespeneKilledEconomy": 0,
            "scoreValueVespeneKilledTechnology": 0,
            "scoreValueVespeneLostArmy": 0,
            "scoreValueVespeneLostEconomy": 0,
            "scoreValueVespeneLostTechnology": 0,
            "scoreValueVespeneUsedActiveForces": 0,
            "scoreValueVespeneUsedCurrentArmy": 0,
            "scoreValueVespeneUsedCurrentEconomy": 0,
            "scoreValueVespeneUsedCurrentTechnology": 0,
            "scoreValueVespeneUsedInProgressArmy": 0,
            "scoreValueVespeneUsedInProgressEconomy": 0,
            "scoreValueVespeneUsedInProgressTechnology": 0,
            "scoreValueWorkersActiveCount": 12
        }
    },
    ...
]
\end{lstlisting}
\end{minipage}

\begin{minipage}{\linewidth}
\begin{lstlisting}[caption={Example UnitTypeChange tracker event JSON object.},language=json,firstnumber=1]
[
    ...
    {
        "evtTypeName": "UnitTypeChange",
        "id": 4,
        "loop": 15,
        "unitTagIndex": 218,
        "unitTagRecycle": 1,
        "unitTypeName": "Egg"
    },
    ...
]
\end{lstlisting}
\end{minipage}

\begin{minipage}{\linewidth}
\begin{lstlisting}[caption={Example UnitBorn tracker event JSON object.},language=json,firstnumber=1]
[
    ...
    {
        "controlPlayerId": 1,
        "evtTypeName": "UnitBorn",
        "id": 1,
        "loop": 652,
        "unitTagIndex": 238,
        "unitTagRecycle": 1,
        "unitTypeName": "Drone",
        "upkeepPlayerId": 1,
        "x": 23,
        "y": 17
    },
    ...
]
\end{lstlisting}
\end{minipage}

\begin{minipage}{\linewidth}
\begin{lstlisting}[caption={Example UnitTypeChange tracker event JSON object.},language=json,firstnumber=1]
[
    ...
    {
        "evtTypeName": "UnitTypeChange",
        "id": 4,
        "loop": 652,
        "unitTagIndex": 203,
        "unitTagRecycle": 1,
        "unitTypeName": "Larva"
    },
    ...
]
\end{lstlisting}
\end{minipage}

\begin{minipage}{\linewidth}
\begin{lstlisting}[caption={Example UnitDied tracker event JSON object.},language=json,firstnumber=1]
[
    ...
    {
        "evtTypeName": "UnitDied",
        "id": 2,
        "killerPlayerId": null,
        "killerUnitTagIndex": null,
        "killerUnitTagRecycle": null,
        "loop": 652,
        "unitTagIndex": 203,
        "unitTagRecycle": 1,
        "x": 23,
        "y": 17
    },
    ...
]
\end{lstlisting}
\end{minipage}

\begin{minipage}{\linewidth}
\begin{lstlisting}[caption={Example UnitPositions tracker event JSON object.},label={lst:UnitPositionsListing},language=json,firstnumber=1]
[
    ...
    {
        "evtTypeName": "UnitPositions",
        "firstUnitIndex": 265,
        "id": 8,
        "items": [
            0,
            50,
            27,
            14,
            49,
            26,
            3,
            49,
            28,
            9,
            48,
            28,
            18,
            50,
            26,
            11,
            50,
            27,
            21,
            55,
            23
        ],
        "loop": 6000
    }
    ...
]
\end{lstlisting}
\end{minipage}

\end{document}